\documentclass[conference]{IEEEtran}

\usepackage[utf8]{inputenc}
\usepackage{amsmath, amssymb}
\usepackage{graphicx}
\usepackage{hyperref}
\usepackage{url}
\usepackage{cite}

\begin{document}

\title{Assessing Simplification Levels in Neural Networks:\\ The Impact of Hyperparameter Configurations on Complexity and Sensitivity}

\author
{
\IEEEauthorblockN{(Joy) Huixin Guan}
\IEEEauthorblockA{Te Herenga Waka—Victoria University of Wellington}
}

\maketitle

\section{Introduction}
This report aims to assess the simplification level of neural networks under different hyperparameter configurations, focusing particularly on two key metrics: Lempel-Ziv complexity and sensitivity. By adjusting activation functions, the number of hidden layers, and the learning rate of the neural network, we analyze how these hyperparameters impact the network's output complexity and sensitivity to input perturbations. The experiment uses the MNIST dataset for a classification task, evaluating how networks configured with various hyperparameters perform in terms of complexity and robustness (as assessed by sensitivity).

The reason for choosing these hyperparameters is their significant influence on the performance of neural networks. The activation function determines the network's non-linear representation capability, and different activation functions can affect how the model responds to inputs~\cite{nwankpa2018activation, kingma2015adam}. The number of hidden layers is directly related to the model's capacity, while the learning rate affects the model's convergence speed and training quality~\cite{kingma2015adam}. By adjusting these parameters, we aim to evaluate their impact on network complexity and robustness.

The complete experimental code and results can be accessed via the Kaggle notebook: \url{https://www.kaggle.com/code/joyhguan/exploring-hyperparameters-in-mnist-networks-compl}.

\section{Theory}
Neural networks exhibit a certain inductive bias, tending to produce simplified output functions rather than complex mapping relationships~\cite{hinton2006reducing}. This means that even when networks are over-parameterized, their outputs still tend to be simple and structured.

To measure the degree of simplification in network outputs, we used the following two metrics in this experiment:

\subsection{Lempel-Ziv Complexity}
This is a computationally feasible approximation of Kolmogorov complexity, used to quantify the complexity of network outputs~\cite{ziv1977universal}. A lower Lempel-Ziv complexity indicates that the output patterns are more simplified and regular~\cite{ziv1977universal, li2008introduction}. In machine learning, lower complexity is often associated with better generalization ability~\cite{lecun2015deep}.

\textbf{Motivation}: The advantage of Lempel-Ziv complexity is that it can be approximated using existing compression algorithms (such as zlib) and effectively represents the structure and redundancy of the data, making it a suitable tool for assessing neural network output complexity. Compared to other complexity calculation methods, Lempel-Ziv complexity is more efficient for evaluating network outputs and is straightforward to compute.

\subsection{Sensitivity}
Sensitivity measures the network's response to input perturbations. By applying small perturbations to the input data and then measuring the magnitude of changes in the network’s output, we can assess the network's robustness. Lower sensitivity means that the network is less affected by minor input changes, indicating a more stable output, which often correlates with greater robustness and resistance to noise~\cite{lecun2015deep}.

\textbf{Motivation}: We chose the L2 norm to calculate the magnitude of output perturbations because the L2 norm intuitively measures the Euclidean distance between two vectors, making it well-suited for evaluating changes in neural network output. Additionally, sensitivity helps us understand how different activation functions and network depths influence the stability of the network's output.

\subsection{Further Explanation of Method Choices}
In this experiment, we chose to use trained neural networks rather than random networks. This is because trained networks better reflect the actual impact of network structure and hyperparameters on their performance. While random networks can eliminate the influence of specific data on the results, they do not effectively demonstrate the learning capability and ability to simplify outputs of a network.

The choice of network configurations (such as different activation functions, the number of hidden layers, and learning rates) is based on their core impact on network performance. The activation function determines the network's non-linear representation capacity, the number of hidden layers is directly linked to the model’s capacity, and the learning rate directly affects the model's convergence speed and training quality~\cite{nwankpa2018activation, kingma2015adam}.

By employing these methods and rationale, we can effectively evaluate how different hyperparameter settings influence the complexity and sensitivity of the network outputs, thus gaining insights into their generalization ability and robustness.

\section{Experiment Setup}
To evaluate the simplification and robustness of the neural networks, we designed seven sets of experiments. Each experiment adjusts the activation function, the number of hidden layers, and the learning rate to assess how these hyperparameters affect the network's complexity and sensitivity. The specific configurations of each experiment are as follows:

\begin{itemize}
    \item \textbf{Index 1}: Activation function = ReLU, Hidden layers = [64, 64], Learning rate = 0.001
    \item \textbf{Index 2}: Activation function = Tanh, Hidden layers = [64, 64], Learning rate = 0.001
    \item \textbf{Index 3}: Activation function = LeakyReLU, Hidden layers = [64, 64, 128], Learning rate = 0.001
    \item \textbf{Index 4}: Activation function = Sigmoid, Hidden layers = [64, 64], Learning rate = 0.001
    \item \textbf{Index 5}: Activation function = ReLU, Hidden layers = [128, 128], Learning rate = 0.001
    \item \textbf{Index 6}: Activation function = ReLU, Hidden layers = [64, 64], Learning rate = 0.1
    \item \textbf{Index 7}: Activation function = ReLU, Hidden layers = [64, 64, 128], Learning rate = 0.001
\end{itemize}

We used the MNIST dataset, which consists of grayscale handwritten digit images with dimensions of $28\times28$ pixels. All images were standardized with a mean of 0.5 and a standard deviation of 0.5. We used the Adam optimizer with learning rates of 0.001 and 0.1, training for 10 epochs. The network complexity and sensitivity were evaluated using the following steps:

\begin{enumerate}
    \item \textbf{Lempel-Ziv Complexity}: We calculated the network's output complexity by compressing the output sequence using the zlib library and measuring the compressed length.
    \item \textbf{Sensitivity}: We applied small perturbations to the input images (epsilon = $1\times10^{-5}$) and calculated the difference in network outputs before and after the perturbation using the L2 norm.
\end{enumerate}

\section{Experimental Results}
\subsection{Training Loss and Accuracy}
\begin{figure}[!t]
\centering
\includegraphics[width=\linewidth]{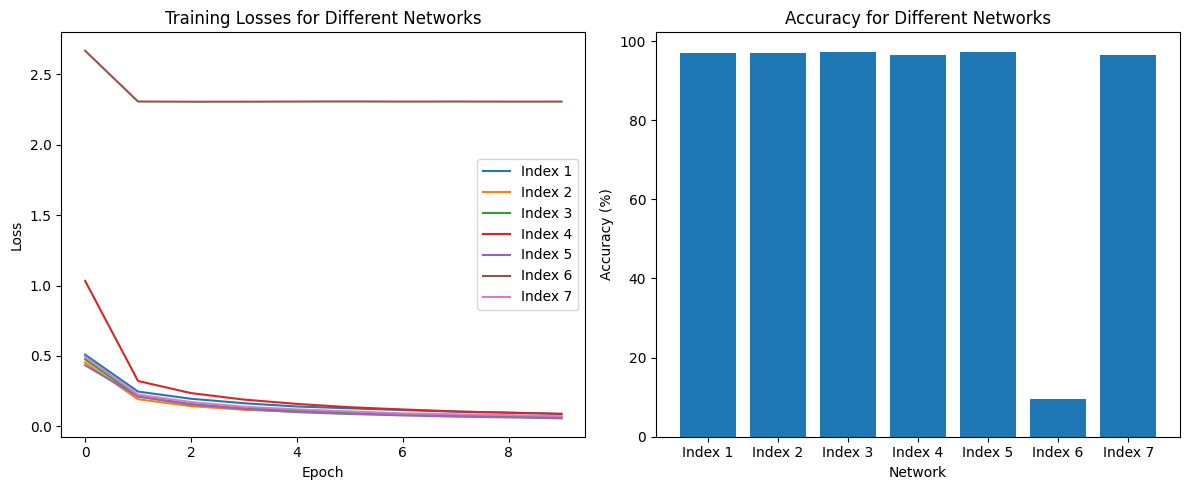}
\caption{training loss and test accuracy}
\label{fig:output1}
\end{figure}

We trained models under seven different configurations and calculated the training loss and test accuracy for each network and presented them in Figure \ref{fig:output1}

Here are the key observations:

\begin{itemize}
    \item \textbf{Indexes 1-5 and 7}: The loss curves gradually decreased and converged, indicating that these networks successfully learned. Their test accuracies were close to 97\%, showing good generalization performance.
    \item \textbf{Index 6} (ReLU with a learning rate of 0.1): The loss did not change much and remained around 2.5, indicating that the network failed to converge. Its test accuracy was only 11.35\%, showing that the learning rate was too high, preventing the network from learning meaningful features.
\end{itemize}

\subsection{Lempel-Ziv Complexity and Sensitivity}
To further compare the simplification and robustness of the network outputs, we calculated the Lempel-Ziv complexity and sensitivity for each network and presented them in Figure \ref{fig:output2}:
\begin{figure}[!t]
\centering
\includegraphics[width=\linewidth]{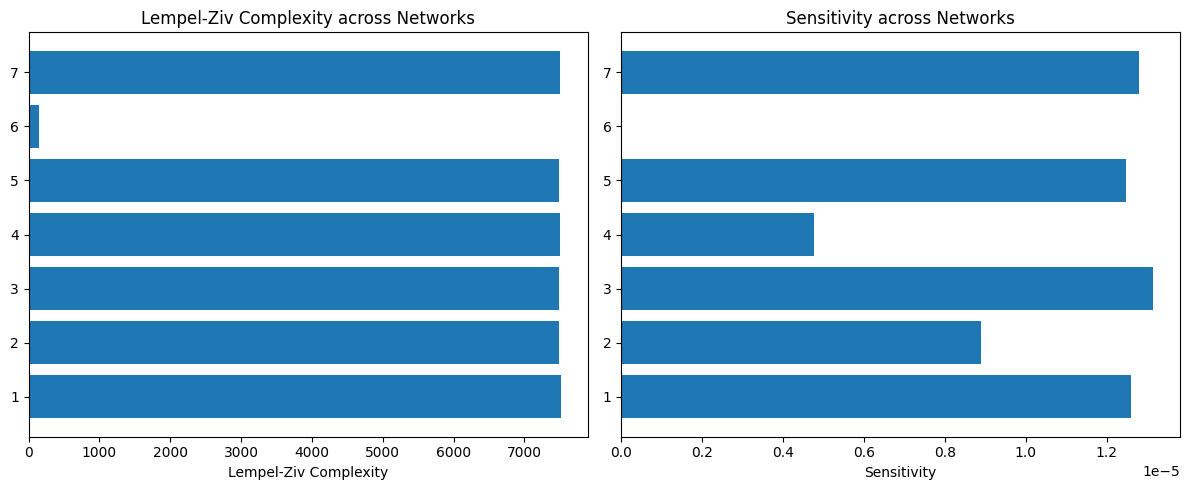}
\caption{complexity and sensitivity}
\label{fig:output2}
\end{figure}

\textbf{Lempel-Ziv Complexity}: Except for Index 6, the Lempel-Ziv complexity of other networks was close to 7500, indicating that these networks produced fairly consistent and complex output patterns. The complexity of Index 6 was significantly lower than that of other networks, demonstrating that the high learning rate prevented the network from learning meaningful features, leading to very simple outputs. This is consistent with Lempel-Ziv complexity as a measure of information compression efficiency~\cite{ziv1977universal, li2008introduction}.

\textbf{Sensitivity}: The choice of activation function had a significant impact on the network's sensitivity. Networks using ReLU and LeakyReLU (e.g., Indexes 3 and 7) showed higher sensitivity, indicating that they responded more strongly to small input changes. This is consistent with the findings of Nwankpa \emph{et al.}~\cite{nwankpa2018activation} who observed that ReLU and LeakyReLU activation functions are more sensitive to input changes in deep networks. Conversely, networks using Sigmoid and Tanh (e.g., Indexes 2 and 4) exhibited lower sensitivity, indicating that these activation functions produce smoother and more stable outputs.

\section{Conclusion}
From this experiment, we arrived at the following conclusions:

\subsection{Impact of Activation Functions}
The choice of activation function significantly affects network sensitivity. Networks using ReLU and LeakyReLU activation functions showed higher sensitivity to input perturbations, while networks using Sigmoid and Tanh activation functions showed lower sensitivity. As observed by Nwankpa \emph{et al.}~\cite{nwankpa2018activation}, this is closely related to the non-linear characteristics of these activation functions. ReLU-type activation functions tend to exhibit "sparsity," responding more rapidly to input changes, while Sigmoid and Tanh tend to produce smoother outputs, reducing their sensitivity to input noise.

\subsection{Impact of Learning Rate}
The choice of learning rate is crucial to the network's training success. A high learning rate (e.g., Index 6, learning rate = 0.1) caused the network to fail to learn meaningful features, as reflected by the low Lempel-Ziv complexity of its output. This is consistent with the findings of Kingma and Ba in their study of the Adam optimizer, where they noted that an excessively high learning rate can lead to unstable training and even failure~\cite{kingma2015adam}. However, the effect of learning rate on network output complexity is more indirect, primarily affecting whether the network successfully learns.

\subsection{Impact of Network Depth}
Increasing the number of hidden layers had a small effect on Lempel-Ziv complexity but, in some cases (e.g., LeakyReLU with 3 layers, Index 7), led to an increase in sensitivity. This suggests that increasing network depth may enhance the network's sensitivity to input perturbations but has little impact on output complexity. This indicates that deeper networks are more likely to affect sensitivity rather than the complexity of output patterns.

Overall, the choice of activation functions and learning rate plays a significant role in determining the model's performance. Future research could explore how these hyperparameters affect network complexity and sensitivity in more complex datasets and tasks.

\section*{Statement}
This report and all related experimental work were conducted independently by Joy Huixin Guan. The tools utilized include Python’s PyTorch library for neural network modeling, the zlib library for calculating Lempel-Ziv complexity, the Matplotlib library for visualization, Overleaf for LaTeX typesetting, and ChatGPT for text polishing and grammar checks. The full code and experimental results can be accessed via the Kaggle platform: \url{https://www.kaggle.com/code/joyhguan/exploring-hyperparameters-in-mnist-networks-compl}.

\bibliographystyle{plain}
\bibliography{references}

\end{document}